\documentclass{article}
\usepackage{spconf,amsmath,graphicx}

\usepackage{enumitem}
\setlist{nosep, leftmargin=14pt}

\usepackage{mwe} 

\title{CELLTRACK R-CNN: A NOVEL END-TO-END DEEP NEURAL NETWORK FOR CELL SEGMENTATION AND TRACKING IN MICROSCOPY IMAGES}
\name{Yuqian Chen$^{1}$, Yang Song$^{2}$, Chaoyi Zhang$^{1}$, Fan Zhang$^{3}$, Lauren O'Donnell$^{3}$}
\address{\emph{Wojciech Chrzanowski$^{4,5}$, Weidong Cai$^{1}$} \\
				\hspace*{\fill} \\
		$^{1}$ School of Computer Science, University of Sydney, Australia \\
		$^{2}$ School of Computer Science and Engineering, University of New South Wales, Australia \\
		$^{3}$ Brigham and Women’s Hospital, Harvard Medical School, USA \\
     $^{4}$ Sydney Pharmacy School, University of Sydney, Australia \\
		$^{5}$ Sydney Nano Institute, University of Sydney, Australia}

\begin{document}
%
\maketitle
\begin{abstract}
Cell segmentation and tracking in microscopy images are of great significance to new discoveries in biology and medicine. In this study, we propose a novel approach to combine cell segmentation and cell tracking into a unified end-to-end deep learning based framework, where cell detection and segmentation are performed with a current instance segmentation pipeline and cell tracking is implemented by integrating Siamese Network with the pipeline. Besides, tracking performance is improved by incorporating spatial information into the network and fusing spatial and visual prediction. Our approach was evaluated on the DeepCell benchmark dataset. Despite being simple and efficient, our method outperforms state-of-the-art algorithms in terms of both cell segmentation and cell tracking accuracies.
\end{abstract}
\section{Introduction}
\label{sec:intro}

Quantitative analysis of microscopy images provides valuable information for understanding cell structures and cell behaviors, which is of great significance in biology and medicine study. To analyze cell behaviors, it is essential for clinical researches to perform cell segmentation and tracking to identify individual cells and follow them over time \cite{C11}. However, the low quality of microscopy images presents special challenges to these tasks.

In recent years, deep learning has proved to be a powerful tool of feature extraction and demonstrated its successful applications in biomedical image analysis. For cell or nuclei segmentation, many studies adopted semantic segmentation architectures such as U-Net to predict foreground and background areas, followed by post-processing procedures \cite{C2}. These methods targeted at capturing details but lacked sufficient object-level information, leading to difficulty in separating individual instances. A few studies used detection based methods and performed segmentation on each detected instance, such as Mask R-CNN \cite{C1}\cite{C3}. These methods demonstrated superior performance in identifying individual instances and no extra post-processing is needed.

The improved detection and segmentation performance of deep learning benefits the subsequent cell tracking task. Most of current tracking methods are conventional algorithms such as overlap intersection-over-union \cite{C4} and Viterbi \cite{C5}, performed independently from detection step. These methods tend to have poor generalization ability and necessitate tuning numerous parameters. Several methods have been proposed to adopt deep learning approaches in cell tracking task [7-13]. In \cite{C12}, a motion model and a classification neural network were combined for cell tracking. \cite{C13} and \cite{C6} attempted to achieve joint cell detection and tracking by predicting cell position likelihood and motion map with a neural network, but they were unable to generate segmentation masks. \cite{C7}, \cite{C14} and \cite{C15} used deep learning techniques for both cell segmentation and tracking but these processes were executed in sequence with two neural networks trained separately. \cite{C8} performed cell segmentation and tracking within a single Recurrent Hourglass Network but showed limited performance and required complex post-processing steps.

In this study, we propose a novel end-to-end neural network framework, dubbed as CellTrack R-CNN, for concurrent cell segmentation and tracking. These tasks are jointly performed by integrating a Siamese tracking branch with the Mask R-CNN pipeline, without any extra post-processing techniques needed. To further enhance tracking performance, spatial information is incorporated into the tracking branch to reach learnable relative position encodings of cell instances, which are then effectively fused with visual features. We evaluated our method on the DeepCell benchmark dataset \cite{C7} with state-of-the-art algorithms, and the results demonstrate superior performance of our CellTrack R-CNN in both cell segmentation and tracking.

\begin{figure*}[t]
     \centering
     \includegraphics[width=15cm]{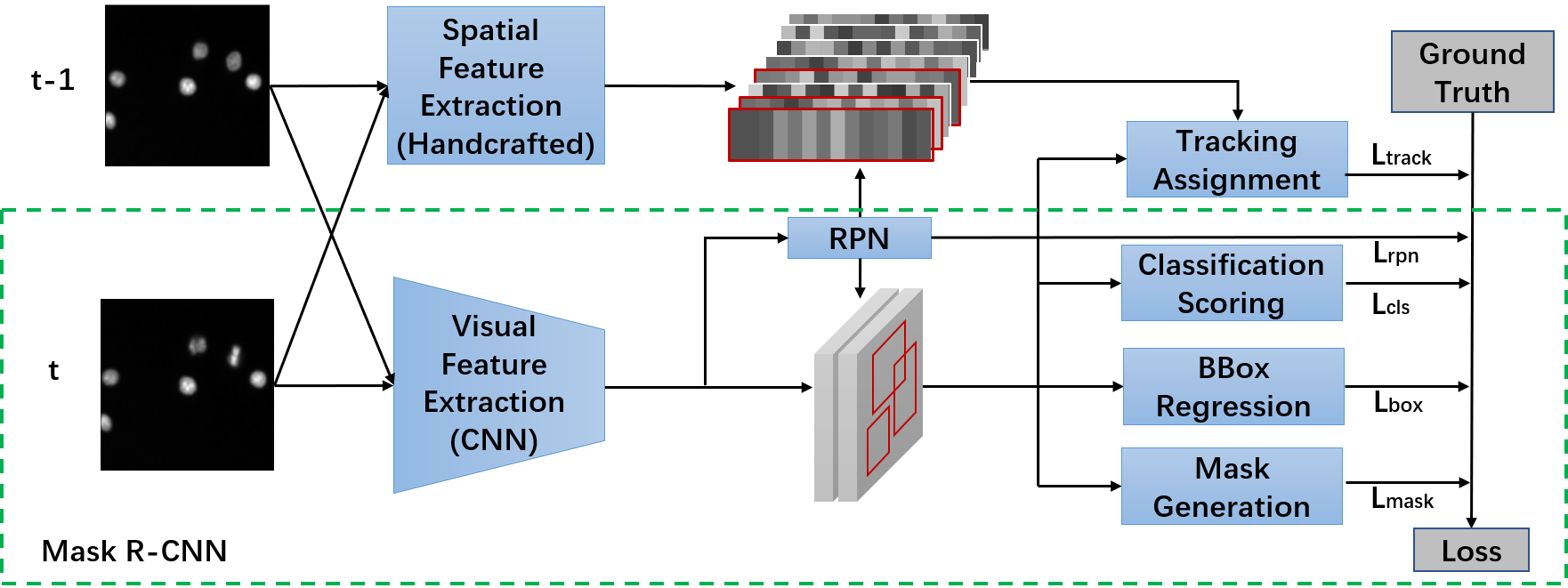}
     \caption{Overall pipeline of our proposed CellTrack R-CNN. Cell detection and segmentation are performed by Mask R-CNN, the architecture in the green box. A Siamese Network branch is integrated into the Mask R-CNN pipeline for tracking. Visual and spatial features of cell instances in two adjacent frames are extracted and put into the tracking head.} \label{fig1}
\end{figure*}
\section{METHODS}
\label{sec:method}

\subsection{Overall Pipeline}
\label{ssec:pipeline}

The overview of our proposed CellTrack R-CNN framework is shown in Fig. \ref{fig1}. We employ Mask R-CNN \cite{C9} as our backbone network for conventional cell detection and segmentation. In addition to the three original branches for predicting classification scores, bounding boxes and segmentation masks, we design a forth branch with a Siamese Neural Network \cite{C10} to achieve cell tracking across frames. Spatial features are extracted in the form of relative position encodings and further aggregated with visual features for better tracking performance. To perform tracking task incorporating temporal information, our framework takes as inputs two adjacent frames to form a training mini-batch.

\subsection{Relative Position Encoding}
\label{ssec:spafea}

To perform tracking task, we need to identify the same and different instances across frames. In the original architecture of Mask R-CNN, only visual features are extracted with convolutional neural networks (CNN). However, in microscopy images, different cells tend to present similar appearance, making visual information insufficient for cell identification.

Therefore, we propose to incorporate spatial information into our CellTrack R-CNN for cell tracking. Inspired by \cite{C16} which studied relative postion encoding in NLP tasks, the spatial feature of each cell is presented as relative position between the target cell and its neighboring ones. Specifically, the coordinate differences in $\mathnormal{x}$ and $\mathnormal{y}$ dimension between the center of target cell and its nearest $\mathnormal{n}$ cells are concatenated in sequence as a vector. In this way, a spatial feature vector with length of 2$\mathnormal{*}$$\mathnormal{n}$ is generated for each cell instance. Parameter $\mathnormal{n}$ is set according to the number of cell instances in images. Spatial vectors are normalized before put into tracking head.

\subsection{Tracking Branch}
\label{ssec:trackbranch}

Similar to the Mask R-CNN framework, in the first stage, a set of candidate boxes is generated from the region proposal network (RPN). Then the proposed box candidates are used to crop visual features and extract spatial features of the corresponding cell instance. In the second stage, we add a tracking head in parallel to the other three branches. Then the obtained visual and spatial features are fed into the tracking head.

\begin{figure}[t]

\begin{minipage}[b]{1.0\linewidth}
  \centering
  \centerline{\includegraphics[width=8.5cm]{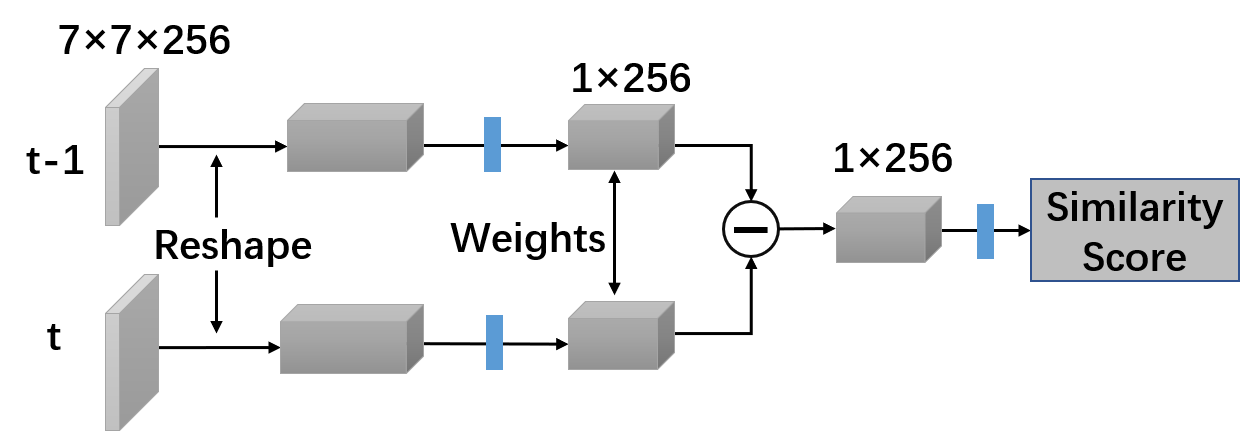}}
  \centerline{(a) Visual tracking head.}\medskip
\end{minipage}
\begin{minipage}[b]{1.0\linewidth}
  \centering
  \centerline{\includegraphics[width=8.5cm]{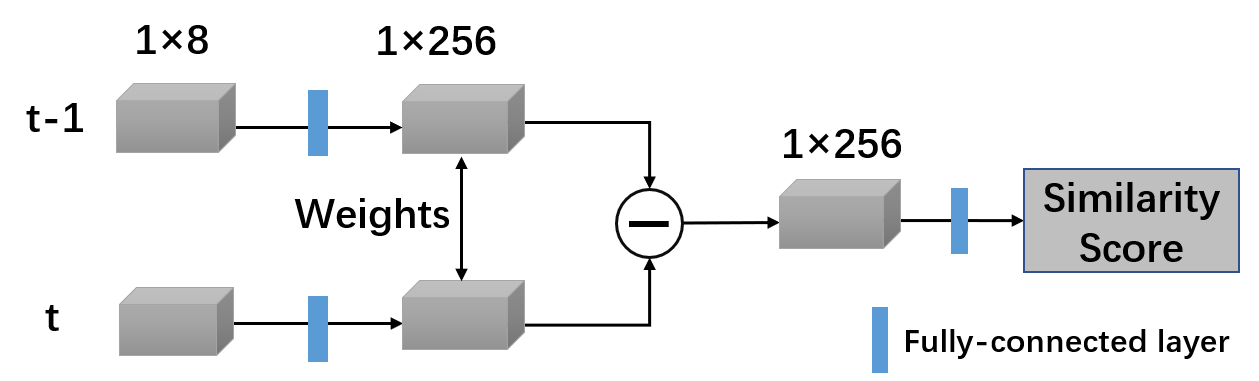}}
  \vspace{0.2cm}
  \centerline{(b) Spatial tracking head.}\medskip
\end{minipage}
\caption{Architecture of tracking branch}
\label{fig2}
\end{figure}

Here we adopt the Siamese Neural Network \cite{C10} in the tracking head. Siamese Network is used to calculate the similarity of two inputs. Two different vectors are put into two neural networks with shared weights and output a similarity score. Our tracking branch is composed of two tracking heads for comparing visual and spatial features respectively, as shown in Fig. \ref{fig2}. In each head, the input vector pair is first transformed to two 256-D vectors with a fully-connected layer, then their difference is calculated and finally a similarity score is obtained with another fully-connected layer.

In the first stage, we have obtained features of candidate proposals in frames $\mathnormal{t}$ and $\mathnormal{t}$-1. For every candidate in frame $\mathnormal{t}$, we calculate its similarity scores with candidates in frame $\mathnormal{t}$-1 respectively with the Siamese tracking head. During the training process, if the candidate pair is instances of the same cell, the label is set to 1, 0 otherwise. The output similarity scores within [0, 1] are used to calculate losses with a cross entropy function. Tracking head is trained in a way that the same instances have scores close to 1 while different ones close to 0. The tracking loss is defined as the average of visual feature loss and spatial feature loss. Our network is trained in an end-to-end fashion. We add the tracking loss into the total losses of our CellTrack R-CNN, which now becomes $\mathnormal{L}$=$L_{rpn}$+$L_{cls}$+$L_{box}$+$L_{mask}$+$L_{track}$.

For inference, each frame in the testing images is processed in sequence with our pipeline. Both mask segmentation and cell tracking are performed on the detected instances. In the tracking task, all instances of the first frame are regarded as new instances of the video and assigned with different tracking ids. To identify the same instance across frames, instances of the current frame are matched to those of the last frame by calculating their visual and spatial similarity scores. After getting the scores, we apply a novel prediction fusion strategy. If the maximum visual and spatial scores correspond to the same cell instance, we assign its tracking id to the target cell. Otherwise, we choose the cell with the maximum overlapping area with the target cell. If the overlapping area is smaller than a threshold $\alpha$, we regard it as a new instance and assign it a new tracking id. If two target cells match to one cell at the last frame, we regard it as cell mitosis. In this way, our method is able to match cell instances across frames and detect appearing, vanishing cells and cell mitosis. After processing all frames, our method generates a set of instance hypothesis and obtains bounding box, classification score, segmentation mask and tracking id for each instance.

\section{EXPERIMENTS AND RESULTS}
\label{sec:experiments}

\subsection{Dataset and Evaluation Metrics}
\label{ssec:data}

We evaluated our method on the public RAW264.7 dataset in the DeepCell database \cite{C7}. It is an annotated dataset specific to live-cell imaging, including fluorescence images of the cell nucleus. This dataset  provides sufficient image sequences which is important for deep learning-based cell tracking algorithms. Besides, all images in this dataset are fully annotated with detection, segmentation and tracking ground truths. The RAW264.7 dataset contains 13 image sequences and there are 30 images in each of them. We split it into three parts with 8 sequences for training, 2 for validation and 3 for testing.

For detection and tracking, we adopted graph-based metric \cite{C11} for evaluation, which was also used in the DeepCell \cite{C7} study. It represents cell lineage in a form of directed acyclic graph (DAG) and generates a tracking score by comparing DAG of generated results and ground truths. Segmentation performance was evaluated by calculating Jaccard Similarity Index of matching objects in generated results and ground truths. These metrics were calculated with publicly available command-line software packages \cite{C11}. We compared our method with state-of-the-art methods using these metrics.

\begin{table}[t]
\caption{Quantitative comparisons of performance on DeepCell dataset. STH: spatial tracking head. D-T: difference between detection and tracking accuracy.}\label{tab1}
~\\
\centering
\setlength{\tabcolsep}{3.5mm}{
\begin{tabular}{ccccc}
\hline
Method & SEG & DET & TRA & D-T\\\hline
DeepCell \cite{C7}& 84.63 & 95.96 & 95.46 & 0.50  \\
FR-Ro-GE \cite{C2}& 81.82 & 95.12 & 94.99 & 0.13\\
Ours w/o STH & 85.22 & 96.93 & 94.19 & 2.74\\
Our method & {\bf 85.26} & {\bf 97.14} & {\bf 97.05} & {\bf 0.09}\\
\hline
\end{tabular}}
\end{table}

\begin{figure}[t]
     \centering
     \includegraphics[width=8.5cm]{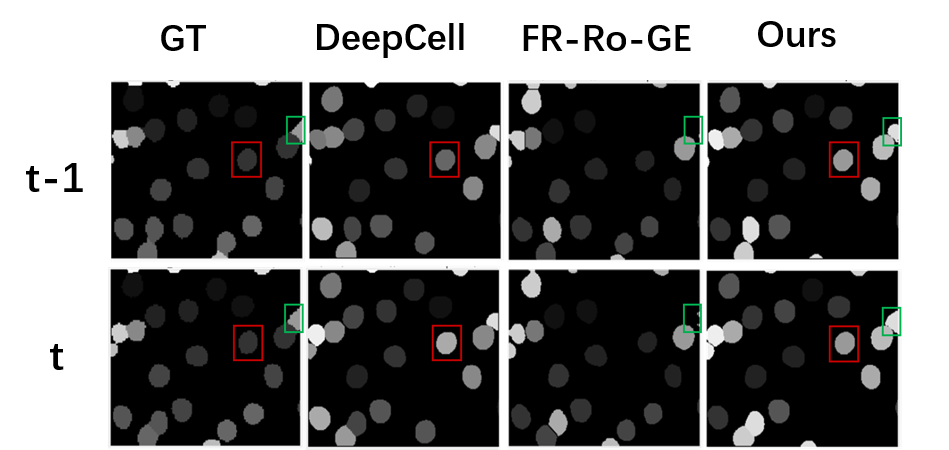}
     \caption{Visualization of experiment results. Red and green boxes indicate wrong trackings of DeepCell and missed detections of FR-Ro-GE, while our method performed correctly.} \label{fig3}
\end{figure}
\subsection{Implementation}
\label{ssec:implem}

We adopted ResNet-50-FPN as the backbone of our CellTrack R-CNN and used the pretrained Mask R-CNN model on MSCOCO Dataset \cite{C17}. Our model was trained end-to-end for 40 epochs. We used SGD optimizer with momentum of 0.9 for training with a learning rate of 0.0025. The original images were resized to (950, 800) as input of the neural network. The hyperparameter $\mathnormal{n}$ in relative position encoding was set to 4 for this dataset because the minimum number of cells in an image was 5. Hyperparameter $\alpha$ during inference was chosen to be 0.1 according to results of the validation set.

\subsection{Results and Discussion}
\label{ssec:result}

In this study, we compared our method with two state-of-the-art studies \cite{C2}\cite{C7}. DeepCell \cite{C7} algorithm is the baseline method of this dataset and FR-Ro-GE \cite{C2} achieves the best tracking performance on the ISBI Cell Tracking Challenge \cite{C11}. In DeepCell, we also used the segmentation pipeline of Mask R-CNN for better comparison of tracking performance.

We calculated the detection, segmentation and tracking accuracies for comparison. Considering that tracking scores were obtained from detection results, the difference between detection and tracking accuracy was also calculated to clarify the tracking performance. As shown in Table \ref{tab1}, our method achieves the best segmentation, detection and tracking performance. Besides, the minimum difference between detection and tracking score further demonstrates the best performance of our cell tracking method. Visualization of cell segmentation and tracking results are shown in Fig. \ref{fig3}.

An important reason that our method outperforms the others is the fusion of spatial information with visual information in the tracking branch. An ablation study was performed to prove the effectiveness of the spatial tracking head. The results in Table 1 indicate that by incorporating spatial information, the tracking performance shows obvious improvement. In fact, cell tracking is specially challenging in microscopy images due to similar appearance of different cells and severe shape deformations over time, making visual information insufficient to identify individual cells across frames. In this situation, spatial information, such as the relative position distribution of neighboring cells, could make great contributions to cell tracking, which has been confirmed by our study. 

In addition to the superior performance, the advantages of our method are its simplicity and efficiency. As we know, most current methods perform cell segmentation and tracking separately and they often require complex and time-consuming post-processing steps. For example, FR-Ro-GE adopted deep learning method only for segmentation and used a separate conventional algorithm for tracking. DeepCell trained two separate neural networks for segmentation and tracking, followed by the conventional tracking algorithm linear programming. However, our method is an end-to-end pipeline. Effective cell tracking is achieved at the price of small increase of model and computation complexity based on an instance segmentation framework.

\section{CONCLUSION}
\label{conclusion}

To our knowledge, CellTrack R-CNN is the first proposed framework that performs concurrent cell segmentation and tracking within a unified neural network without any post-processing needed. These tasks are fulfilled by integrating Siamese Network with Mask R-CNN pipeline. Spatial information represented as relative position encodings of cell instances is incorporated into the network and proves great benefits to the tracking performance. Our method shows state-of-the-art performance in terms of cell segmentation and tracking. In the future, we will explore more effective strategies for fusing visual and spatial information and investigate performance of our method under various clinical situations.

\section{Compliance with Ethical Standards}
\label{sec:ethics}

No ethical approval was required for using public dataset.

\section{Acknowledgments}
\label{sec:acknowledgments}

No funding was received for this study. The authors have no relevant financial or non-financial interests to disclose.

\bibliographystyle{IEEEbib}
\bibliography{yche_ISBI_latex}

\renewcommand{\baselinestretch}{1.1}
\begin{thebibliography}{10}
\footnotesize
\addtolength{\itemsep}{-0.6 em}


\bibitem{C11}
M.~Ma{\v{s}}ka, V.~Ulman, D.~Svoboda, et~al.,
\newblock ``A benchmark for comparison of cell tracking algorithms,''
\newblock {\em Bioinformatics}, vol. 30, no. 11, pp. 1609--1617, 2014.

\bibitem{C2}
O.~Ronneberger, P.~Fischer, and T.~Brox,
\newblock ``U-net: Convolutional networks for biomedical image segmentation,''
\newblock in {\em MICCAI}. Springer, 2015, pp. 234--241.

\bibitem{C1}
H.~Tsai, J.~Gajda, T.~Sloan, et~al.,
\newblock ``Usiigaci: Instance-aware cell tracking in stain-free phase contrast
  microscopy enabled by machine learning,''
\newblock {\em SoftwareX}, vol. 9, pp. 230--237, 2019.

\bibitem{C3}
D.~Liu, D.~Zhang, Y.~Song, C.~Zhang, F.~Zhang, L.~O'Donnell, and W.~Cai,
\newblock ``Nuclei segmentation via a deep panoptic model with semantic feature
  fusion,''
\newblock in {\em IJCAI}, 2019, pp. 861--868.

\bibitem{C4}
E.~Bochinski, V.~Eiselein, and T.~Sikora,
\newblock ``High-speed tracking-by-detection without using image information,''
\newblock in {\em AVSS}. IEEE, 2017, pp. 1--6.

\bibitem{C5}
K.~Magnusson, J.~Jald{\'e}n, P.~M Gilbert, and H.~Blau,
\newblock ``Global linking of cell tracks using the viterbi algorithm,''
\newblock {\em IEEE transactions on medical imaging}, vol. 34, no. 4, pp.
  911--929, 2014.

\bibitem{C12}
T.~He, H.~Mao, J.~Guo, and Z.~Yi,
\newblock ``Cell tracking using deep neural networks with multi-task
  learning,''
\newblock {\em Image and Vision Computing}, vol. 60, pp. 142--153, 2017.

\bibitem{C13}
J.~Hayashida and R.~Bise,
\newblock ``Cell tracking with deep learning for cell detection and motion
  estimation in low-frame-rate,''
\newblock in {\em MICCAI}. Springer, 2019, pp. 397--405.

\bibitem{C6}
J.~Hayashida, K.~Nishimura, and R.~Bise,
\newblock ``Mpm: Joint representation of motion and position map for cell
  tracking,''
\newblock in {\em CVPR}, 2020, pp. 3823--3832.

\bibitem{C7}
E.~Moen, E.~Borba, G.~Miller, et~al.,
\newblock ``Accurate cell tracking and lineage construction in live-cell
  imaging experiments with deep learning,''
\newblock {\em bioRxiv}, p. 803205, 2019.

\bibitem{C14}
J.~Lugagne, H.~Lin, and M.~Dunlop,
\newblock ``Delta: Automated cell segmentation, tracking, and lineage
  reconstruction using deep learning,''
\newblock {\em PLoS computational biology}, vol. 16, no. 4, pp. e1007673, 2020.

\bibitem{C15}
A.~Panteli, D.~Gupta, N.~de~Bruin, and E.~Gavves,
\newblock ``Siamese tracking of cell behaviour patterns,''
\newblock in {\em MIDL}, 2020.

\bibitem{C8}
C.~Payer, D.~{\v{S}}tern, T.~Neff, et~al.,
\newblock ``Instance segmentation and tracking with cosine embeddings and
  recurrent hourglass networks,''
\newblock in {\em MICCAI}. Springer, 2018, pp. 3--11.

\bibitem{C9}
K.~He, G.~Gkioxari, P.~Doll{\'a}r, and R.~Girshick,
\newblock ``Mask r-cnn,''
\newblock in {\em ICCV}, 2017, pp. 2961--2969.

\bibitem{C10}
S.~Chopra, R.~Hadsell, and Y.~LeCun,
\newblock ``Learning a similarity metric discriminatively, with application to
  face verification,''
\newblock in {\em CVPR'05}. IEEE, 2005, vol.~1, pp. 539--546.

\bibitem{C16}
P.~Shaw, J.~Uszkoreit, and A.~Vaswani,
\newblock ``Self-attention with relative position representations,''
\newblock in {\em NAACL-HLT (2)}, 2018.

\bibitem{C17}
T.~Lin, M.~Maire, S.~Belongie, et~al.,
\newblock ``Microsoft coco: Common objects in context,''
\newblock in {\em ECCV}. Springer, 2014, pp. 740--755.

\end{thebibliography}

\end{document}